\documentclass{article}

\usepackage{microtype}
\usepackage{graphicx}
\usepackage{subfigure}
\usepackage{booktabs} 

\usepackage{hyperref}

\usepackage[accepted]{icml2024}

\usepackage{amsmath}
\usepackage{amssymb}
\usepackage{mathtools}
\usepackage{amsthm}

\usepackage[capitalize,noabbrev]{cleveref}

\theoremstyle{plain}

\theoremstyle{definition}

\theoremstyle{remark}

\usepackage[textsize=tiny]{todonotes}

\icmltitlerunning{Hybrid Recurrent Models Support Emergent Descriptions for Hierarchical Planning and Control}
\allowdisplaybreaks

\begin{document}

\twocolumn[\icmltitle{Hybrid Recurrent Models Support Emergent Descriptions for Hierarchical Planning and Control}



\icmlsetsymbol{equal}{*}

\begin{icmlauthorlist}
\icmlauthor{Poppy Collis}{equal,sussex}
\icmlauthor{Ryan Singh}{equal,sussex,verses}
\icmlauthor{Paul F Kinghorn}{sussex}
\icmlauthor{Christopher L Buckley}{sussex,verses}
\end{icmlauthorlist}

\icmlaffiliation{sussex}{School of Engineering and Informatics, University of Sussex, Brighton, UK}
\icmlaffiliation{verses}{VERSES AI Research Lab, Los Angeles, California, USA}

\icmlcorrespondingauthor{Poppy Collis}{pzc20@sussex.ac.uk}

\icmlkeywords{Hybrid control, Piecewise affine approximations, Exploration, Decision-making}
\vskip 0.3in
]


\printAffiliationsAndNotice{\icmlEqualContribution} 


\begin{abstract}
An open problem in artificial intelligence is how systems can flexibly learn discrete abstractions that are useful for solving inherently continuous problems. Previous work has demonstrated that a class of hybrid state-space model known as recurrent switching linear dynamical systems (rSLDS) discover meaningful behavioural units via the piecewise linear decomposition of complex continuous dynamics \cite{lindermanRecurrentSwitchingLinear2016}. Furthermore, they model how the underlying continuous states drive these discrete mode switches. We propose that the rich representations formed by an rSLDS can provide useful abstractions for planning and control. We present a novel hierarchical model-based algorithm inspired by Active Inference in which a discrete MDP sits above a low-level linear-quadratic controller. The recurrent transition dynamics learned by the rSLDS allow us to (1) specify temporally-abstracted sub-goals in a method reminiscent of the options framework, (2) lift the exploration into discrete space allowing us to exploit information-theoretic exploration bonuses and (3) `cache' the approximate solutions to low-level problems in the discrete planner. We successfully apply our model to the sparse Continuous Mountain Car task, demonstrating fast system identification via enhanced exploration and non-trivial planning through the delineation of abstract sub-goals.
\end{abstract}

\begin{figure*}[t]
    \vskip 0.2in
    \begin{center}
        \centerline{\includegraphics[width=0.8\textwidth]{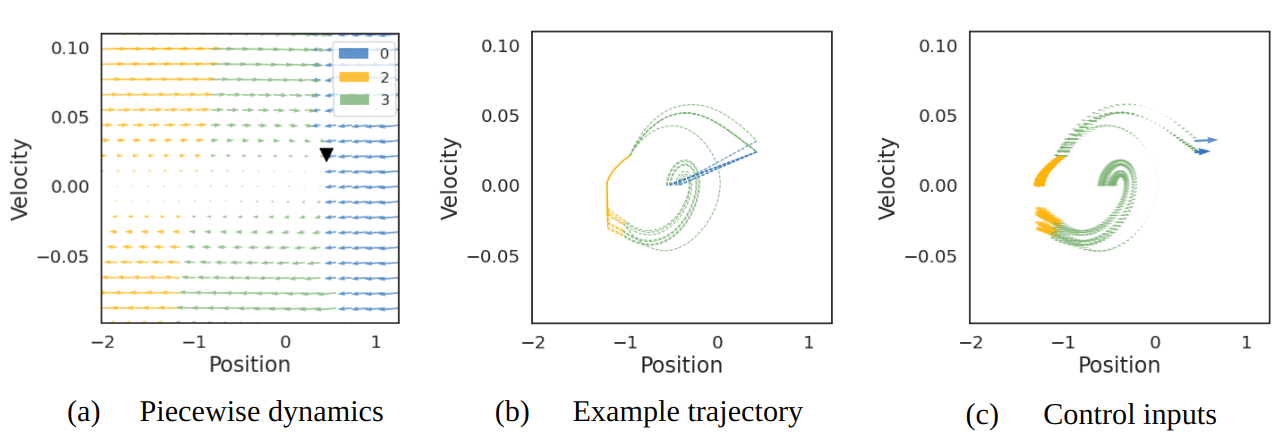}}
        \caption{\textbf{HHA solves nonlinear problems via specifying abstract sub-goals in state-space.} (a) Piecewise linear dynamics  of the Continuous Mountain Car state-space found by rSLDS. Reward location shown (\emph{black triangle}). While the rSLDS retrives 5 modes in total, here we plot only the modes seen in the position-velocity ($x$) space without showing the control input ($u$) axis. (b) Example trajectory (segments coloured by mode) showing the HHA consistently navigating to the goal. (c) Continuous control input (coloured by discrete action specified by planner and arrow size indicating magnitude and direction) over same example trajectory in (b).}
        \label{piecewise}
    \end{center}
    \vskip -0.2in
\end{figure*}

\section{Introduction}
In a world that is inherently continuous, the brain’s apparent capacity to distil discrete concepts from sensory data represents a highly desirable feature in the design of autonomous systems. Humans are able to flexibly specify abstract sub-goals during planning, thereby reducing problems into manageable chunks \cite{newell1972human, Gobet2001}. Furthermore, they are able to transfer this knowledge across new tasks; a process which has proven a central challenge in artificial intelligence \cite{Garcez2023}. Translating problems into discrete space offers distinct advantages in decision-making. Namely, the computationally feasible application of information-theoretic measures (e.g. information-gain), as well as the direct implementation of classical techniques such as dynamic programming \cite{LaValle_2006,fristonstructure}. One prevalent approach to tackling continuous spaces involves the simple grid-based discretisation of the state-space, however this becomes extremely costly as the dimensionality increases \cite{coulom2007, Mnih2015HumanlevelCT}. We therefore ask how we might be able to smoothly handle the presence of continuous variables whilst maintaining the benefits of decision-making in the discrete domain.

To address this, we explore the rich representations learned by recurrent switching linear dynamical systems (rSLDS) in the context of planning and control. This class of hybrid state-space model consists of discrete latent states that evolve via Markovian transitions, which act to index a discrete set of linear dynamical systems \cite{lindermanRecurrentSwitchingLinear2016}. Importantly, a continuous dependency in the discrete state transition probabilities is included in the generative model. By providing an understanding of the continuous latent causes of switches between discrete modes, this recurrent transition structure can be exploited such that a controller can flexibly specify inputs to drive the system into a desired  region of the state-action space. By embracing the established control-theoretic strategy of piecewise linear decomposition of  nonlinear dynamics, our approach lies in contrast to the comparatively opaque solutions found by  continuous function approximators \cite{liberzon2003switching,Mnih2015HumanlevelCT}. Using statistical methods to fit these models provides a means by which we can effectively perform online discovery of useful non-grid discretisations of the state-space for system identification and control.

We describe a novel model-based algorithm inspired by Active Inference \cite{parr2022active}, in which a discrete MDP, informed by the representations of an rSLDS, interfaces with a finite horizon linear-quadratic regulator (LQR) implementing closed-loop control. We demonstrate the efficacy of this algorithm by applying it to the classic control task of Continuous Mountain Car \cite{openai_continuous_mountain_car}. We show that information-theoretic exploration drive integrated with the emergent piecewise description of the task-space facilitates fast system identification to find successful solutions to this non-trivial planning problem.

\subsection{Contributions}
\begin{itemize}
    \item The enhancement of planning via the introduction of temporally-abstracted sub-goals by decoupling a discrete MDP from the continuous clock time using the emergent representations from an rSLDS.
    \item The lifting of information-seeking decision-making into a (discrete) abstraction of the states enabling efficient exploration and thereby reducing sensitivity to the dimensionality of the task-space.
\end{itemize}

\section{Related work}
In the context of control, hybrid models in the form of piecewise affine (PWA) systems have been rigorously examined and are widely applied in real-world scenarios \cite{Bemporad2000,Borrelli2006}. Previous work by Abdulsamad et. al. has applied a variant on rSLDS (recurrent autoregressive hidden Markov models) to the optimal control of general nonlinear systems \cite{10128705, pmlr-v120-abdulsamad20a}. The authors use these models to the approximate expert controllers in a closed-loop behavioural cloning context. While their algorithm focuses on value function approximation, in contrast, we learn online without expert data and focus on flexible discrete planning.

\begin{figure*}[t]
    \vskip 0.2in
    \begin{center}
        \centerline{\includegraphics[width=0.8\textwidth]{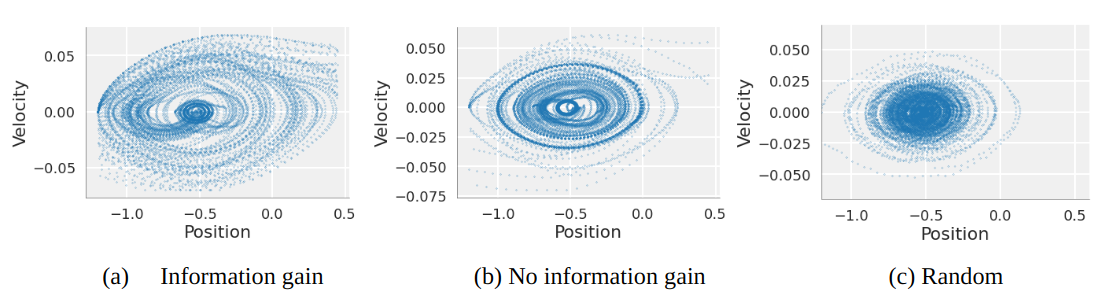}}
        \caption{\textbf{HHA with information-gain explored a wider range of the state-space}. State-space coverage in Continuous Mountain Car after 10,000 steps and best of 3 runs for (a) HHA with information-gain drive, (b) HHA without information gain drive and (c) randomly sampled continuous actions baseline.}
        \label{fig:exploration}
    \end{center}
    \vskip -0.2in
\end{figure*}

\section{Framework}
Here, we provide a overall outline of the approach to approximate control taken with our Hybrid Hierarchical Agent (HHA) algorithm. Consider that we have decomposed the nonlinear dynamics into piecewise affine regions of the state-space using an rSLDS. Should the HHA wish to navigate to a goal specified in continuous space, the recurrent generative model parameters of the rSLDS allow it to identify the discrete region within which the goal resides, thereby lifting the goal into a high-level objective. The agent may then generate a plan at a discrete level, making use of the information-seeking bonuses that this affords. Planning translates to specifying a sequence of abstract sub-goals. Again using the recurrent generative model, the agent can specify, for each sub-goal region, a continuous point in state-space with which to drive the system into. Once in the discrete goal region, the agent straightforwardly navigates to the continuous goal. The following sections detail the components of the HHA. For additional information, please refer to Appendix.~\ref{appendix}

\subsection{rSLDS(ro)}
In the recurrent-only (ro) formulation of the rSLDS, the discrete latent states $z_t \in \{1, 2, . . . , K\}$ are generated as a function of the continuous latents $x_t \in \mathbb R^M$ and the control input $u_t \in \mathbb R^N$ via a softmax regression model
\begin{align}
    P(z_{t+1}|x_t, u_t) = softmax(W_x x_t + W_u u_t + r)
\end{align}
whereby $W_x$ and $W_u$ are weight matrices with dimensions $\mathbb R^{K \times M}$ and $r$ is a bias of size $\mathbb R^{K}$. The continuous dynamics evolve according to a discrete linear dynamical system indexed by $z_t$ with Gaussian diagonal noise,
\begin{equation}
\begin{split}
    x_{t+1}|x_t, u_t, z_t &= A_{z_{t}} x_t + B_{z_{t}} u_t + b_{z_t} + \nu_t, \\
    &\nu_t \sim \mathcal{N}(0, Q_{z_{t}})
\end{split}
\end{equation}
\begin{equation}
    y_t| x_t = C_{z_{t}}x_t + \omega_t, \;\; \omega_t \sim \mathcal{N}(0, S_{z_{t}})
\end{equation}
and identity emissions model with Gaussian diagonal noise. In order to learn the rSLDS parameters using Bayesian updates, conjugate matrix normal inverse Wishart (MNIW) priors are placed on the parameters of the dynamical system and recurrence weights. Inference requires approximate methods given that the recurrent connections break conjugacy rendering the conditional likelihoods non-Gaussian. Details of the Laplace Variational Expectation Maximisation algorithm used is detailed in \cite{zoltowski2020unifying}.

\subsection{Discrete planner}

We have a Bayesian Markov Decision Process (MDP) \cite{Vlassis2012} described by $\mathcal{M}_B = (S, A, P_a, R, P_\theta)$. $S$ represents the set of all possible discrete states of the system and are essentially a re-description of the discrete latents $Z$ found by the rSLDS. $A$ is the set of all possible actions which, in our case, is equal to the number of states $S$. The state transition probabilities, $p_a(s_{t+1}\mid s_t=s, a_t=a, \theta)\sim Cat(\theta_{as})$,
and are parameterised by $\theta \in \mathbb{R}^{s\times s \times a}$ for which we maintain Dirichlet priors over, $p(\theta_{as}) \sim Dir(\alpha_{as})$, facilitating directed exploration. Due to conjugate structure, as the agent obtains new empirical information, Bayesian updates amount to a simple count-based update of the Dirichlet parameters \cite{murphy2012machine}. Importantly, the structure of the state transition model has been constrained by the adjacency structure of the polyhedral partitions extracted from recurrent transition dynamics of the rSLDS: invalid transitions are assigned zero probability while valid transitions are assigned a high probability (see \ref{extract-adj}). $R$ is the reward function which, translated into the Active Inference framework, acts as a prior distribution over rewarding states providing the agent with an optimistic bias during policy inference \cite{millidge2020relationship,parr2022active}. 

The discrete planner outputs a discrete action, where the first action is taken from a receding horizon optimisation:
\begin{align}
    a_0 &= \arg \min_{a_{1:T}} J(a_{1:T}) \\
    J(a_{1:T}) &= \mathbb{E}[\sum_{t=0}^{T} R(s_t, a_t) + IG_{t}(\alpha)  \mid s_0, {a_{1:T}}].
\end{align}
This includes an explicit information-seeking incentive $IG_t(\alpha)$ (see \ref{online-problem}).
This descending discrete action $a_0$ is translated into a continuous control prior $x_j$ via the following link function,
\begin{align}
x_j = \underset{x}{\arg\max} \, P(z=j|x, u)
\end{align}
which represents an approximately central point in the desired discrete region $j$ requested by action $a_0$ (see \ref{control-priors}). The ascending messages from the continuous level are translated into a categorical distribution via the rSLDS softmax link function. Importantly, the discrete planner is only triggered when the system switches into a new mode \footnote{Or a maximum dwell-time (hyperparameter) is reached.}. In this sense, discrete actions are temporally abstracted and decoupled from continuous clock-time in a method reminiscent of the options framework \cite{SUTTON1999181, Daniel2016}.
\subsection{Continuous controller}
\label{continuous-controller}
Continuous closed-loop control is handled by a finite-horizon linear-quadratic regulator (LQR) controller. For controlling the transition from mode $i$ to mode $j$ ($x_i$ to $x_j$). The objective of the LQR controller is to minimise the following quadratic cost function:
\begin{align}
    \pi_{ij}(x) &= \arg \min_{\pi} J_{ij}(\pi) \\
    J_{ij}(\pi) &= 
    \mathbb{E}_{\pi, x_i}\bigg[(x_S - x_j)^T Q_f (x_S - x_j) \notag \\
    &\quad + \sum_{t=0}^{S-1} u_t^T R u_t\bigg]
\end{align}
where $S$ is the finite time horizon, $Q_f$ is the matrix that penalises the terminal state deviation from $x_j$ and $R$ is the control cost where high control input is penalised such that the controller only provides solutions within constraints (for further discussion, see Sec.~\ref{discussion}). The approximate closed-loop solution to each of these sub-problems is computed offline by taking in the parameters of the linear systems indexed by the discrete modes and the continuous control priors acting as reference points (see \ref{LQR}). 

\section{Results}
\label{methods}

To evaluate the performance of our (HHA) model, we applied it to the classic control problem of Continuous Mountain Car. This problem is particularly relevant for our purposes due to the sparse nature of the rewards, necessitating effective exploration strategies to achieve good performance. The HHA is initialised according to the procedure outlined in \cite{lindermanRecurrentSwitchingLinear2016}. The rSLDS parameters are then fitted to the observed trajectories every 1000 steps of the environment unless a reward threshold within a single episode is reached.

\begin{figure}[htbp]
    \centering
    \includegraphics[width=0.6\linewidth]{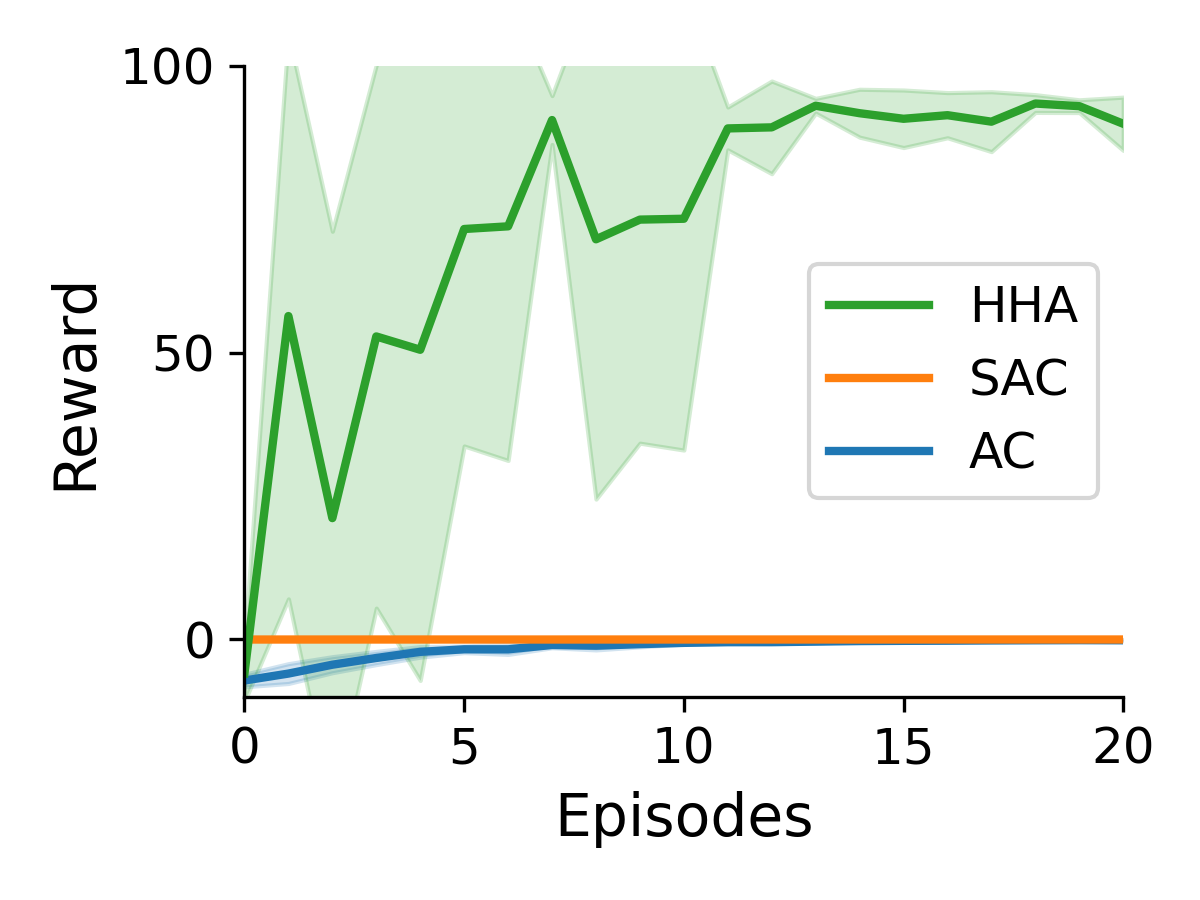}
    \caption{\textbf{HHA both finds the reward and captilises on its experience significantly quicker than other model-free RL baselines}. Average reward (+/- std) over 6 runs for Continuous Mountain Car (20 episodes, max episode length of 200 steps) for HHA (our model), Soft-Actor Critic (with 2 Q-functions), and Actor-Critic models. Note that after 20 episodes, SAC and AC are yet to find the reward and converge on a solution.} \label{fig:rewards}
\end{figure}

We find that the HHA finds piecewise affine approximations of the task-space and uses these discrete modes effectively to solve the task. Fig.\ref{piecewise} shows that while the rSLDS has divided up the space according to position, velocity and control input, the useful modes for solving the task are those found in the position space. Once the goal and a good approximation to the system has been found, the HHA successfully and consistently navigates to the reward.

Fig.~\ref{fig:exploration} shows that the HHA performs a comprehensive exploration of the state-space and significant gains in the state-space coverage are observed when using information-gain drive in policy selection compared to without. Interestingly, even without information-gain, the area covered by the HHA is still notably better than that of the random action control. This is because the non-grid discretisation of the state-space significantly reduces the dimensionality of the search space in a behaviourally relevant way.

We compare the performance of the HHA to other reinforcement learning baselines (Actor-Critic and Soft Actor-Critic) and find that the HHA both finds the reward and captilises on its experience significantly quicker than the other models (see Fig.~\ref{fig:rewards}). Indeed, our model competes with the state-space coverage achieved by model-based algorithms with exploratory enhancements in the discrete Mountain Car task, which is inherently easier to solve (see ~\ref{DQNMBE}).

\section{Discussion}
\label{discussion}
Through the application of our Hybrid Hierarchical Agent to the Continuous Mountain Car problem, we have demonstrated that rSLDS representations hold promise for enriching planning and control. The emergence of non-grid discretisations of the state-space allows us to perform fast systems identification via enhanced exploration, and successful non-trivial planning through the delineation of abstract sub-goals. Hence, the time spent exploring each region is not equivalent in euclidean space which helps mitigate the curse of dimensionality that other grid-based methods suffer from. 

Such a piecewise affine approximation of the space will incur some loss of optimality in the long run when pitted against black-box approximators. This is due to the nature of caching only approximate closed-loop solutions to control within each piecewise region, whilst the discrete planner implements open-loop control. However, this approach eases the online computational burden for flexible re-planning. Hence in the presence of noise or perturbations within a region, the controller may adapt without any new computation. This is in contrast to other nonlinear model-based algorithms like  model-predictive control where reacting to disturbances requires expensive trajectory optimisation at every step \cite{Schwenzer2021}. By using the piecewise affine framework, we maintain functional simplicity and interpretability through structured representation. This method is amenable to future alignment with a control-theoretic approach to safety guarantees for ensuring robust system performance and reliability.

We acknowledge there may be better solutions to dealing with control input constraints than the one given in Sec.~\ref{continuous-controller}. Different approaches have been taken to the problem of implementing constrained-LQR control, such as further piecewise approximation based on defining reachability regions for the controller \cite{BEMPORAD20023}.

\newpage

\section*{Impact Statement}

This paper presents work whose goal is to advance the field of 
Machine Learning. There are many potential societal consequences 
of our work, none which we feel must be specifically highlighted here.

\bibliographystyle{icml2024}
\bibliography{example-paper}

\newpage
\appendix
\onecolumn
\appendix

\section{Appendix / supplemental material}
\label{appendix}

\subsection{Framework}
\label{app:framework}

\textbf{Optimal Control} 

We adopt the optimal control framework, 
specifically we consider discrete time state space dynamics of the form:
\begin{equation}
    x_{t+1} = f(x_t, u_t, \eta_t)
\end{equation}
with known initial condition $x_0$, and $\eta_t$ drawn from some time invariant distribution $\eta_t \sim D$, where $f$ we  assume $p(x_{t+1} \mid x_t, u_t)$ is a valid probability density throughout.

We use  $c_t: X \times U \rightarrow  \mathbb{R}$ for the control cost function at time $t$ and let $\mathbb{U}$ be the set of admissible (non-anticipative, continuous) feedback control laws, possibly restricted by affine constraints. The optimal control law for the finite horizon problem is given as: 
\begin{align}
    J(\pi) &= \mathbb{E}_{x_{0},\pi}[\sum_{t=0}^{T} c_t(x_t, u_t)] \\
    \pi^{*} &= \arg \min_{\pi \in \mathbb{U}} J(\pi) 
\end{align}

\textbf{PWA Optimal Control}

The fact we do not have access to the true dynamical system $f$ motivates the use of a piecewise affine (PWA) approximation. Also known as hybrid systems:
\begin{align}
    x_{t+1} &= A_{i} x_t + B_{i} u_t + \epsilon_t \\
    & \text{when } (x_t, u_t) \in H_i
\end{align}
Where $\mathbb{H}=\{H_i: i \in [K] \}$ is a polyhedral partition of the space $X\times U$.

In the case of a quadratic cost function, it can be shown the optimal control law for such a system is peicewise linear. Further there exist many completeness (universal approximation) type theorems for peicewise linear approximations implying if the original system is controllable, there will exist a peicewise affine approximation through which the system is still controllable \cite{Bemporad2000, Borrelli2006}.

\textbf{Relationship to rSLDS(ro)}

We perform a canonical decomposition of the control objective $J$ in terms of the components or modes of the system. By slight abuse of notation $[x_t = i]:=[(x_t, u_t) \in H_i]$ represent the Iverson bracket.
\begin{align}
    J(\pi) &=  \sum_t \int p_{\pi}(x_t \mid x_{t-1}, u_t)c_t(x_t, u_t) d x_{t} dx_{t-1} \\
    &= \sum_t \int \sum_{i\in [K]} [x_{t-1} = i]p_{\pi}(x_t \mid x_{t-1}, u_t)c_t(x_t, u_t) d x_{t} dx_{t-1} \\
\end{align}
Now let $z_t$ be the random variable on $[K]$ induced by $Z_t = i$ if $[x_t = i]$ we can rewrite the above more concisely as,
\begin{align}
    J(\pi) &=\sum_t \int \sum_{i\in [K]} p_{\pi}(x_t, z_{t-1}=i \mid x_{t-1}, u_t)c_t(x_t, u_t) d x_{t} dx_{t-1} \\
    &= \sum_{i\in [K]}\sum_t \int  p_{\pi}(x_t, z_{t-1}=i \mid x_{t-1}, u_t)c_t(x_t, u_t) d x_{t} dx_{t-1} \\
    &=\sum_{i\in [K]}\sum_t \mathbb{E}_{\pi_i}[c_t(x_t, u_t)] \\
\end{align}
which is just the expectation under a recurrent dynamical system with deterministic switches. Later (see \ref{online-problem}), we exploit the non-deterministic switches of rSLDS in order to drive exploration.

\subsection{Hierarchical Decomposition}

Our aim was to decouple the discrete planning problem from the fast low-level controller. In order to break down the control objective in this manner, we first create a new discrete variable which simply tracks the transitions of $z$, this allows the discrete planner to function in a temporally abstracted manner.

\textbf{Decoupling from clock time} Let the random variable $(\zeta_s)_{s>0}$ record the transitions of $(z_t)_{t>0}$ i.e. let 
\begin{equation}
    \tau_s(\tau_{s-1})= \min \{t: z_{t+1} \neq z_t,  t> \tau_{s-1}\}, \tau_0=0
\end{equation}
be the sequence of first exit times, then $\zeta$ is given by $\zeta_s = z_{\tau_s}$.

With these variables in hand, we frame a small section of the global problem as a first exit problem.

\textbf{Low level problem} Consider the first exit problem defined by, 
\begin{align}
    \pi_{ij}(x_0) &= \arg \min_{\pi, S} J_{ij}(\pi, x_0, S) \\
    J_{ij}(\pi, x_0, S)&= \mathbb{E}_{\pi, x_0}[\sum_{t=0}^{S}c(x_t, u_t)]\\
    & \text{s.t. } (x_t, u_t) \in H_i \\
    & \text{s.t. } c(x, u) = 0 \text{ when } (x,u) \in \partial H_{ij}
\end{align}
where $\partial H_{ij}$ is the boundary $H_i \bigcap H_j$.

Due, to convexity of the polyhedral partition, the full objective admits the decomposition into subproblems
\begin{align}
    J(\pi) &= \sum_s J_{\zeta(s+1), \zeta(s)}(\pi) \\
\end{align}

\textbf{Slow and fast modes} The goal is to tackle the decomposed objectives individually, however the hidden constraint that the trajectories line up presents a computational challenge. Here we make the assumption that the difference in cost induced by different starting positions, induces a relatively small change in the minimum cost $J_{ij}$, intuitively this happens if the minimum state cost in each mode is relatively uniform as compared to the difference between regions.

\textbf{High level problem} If the above assumption holds, we let $J_{ij}^* = \min_{\pi} \int_{x_0}J_{ij}(\pi, x_0)p(x_0)$ be the average cost of each low-level problem. We form a markov chain:
\begin{equation}
    p_{ik}(u) = \mathbb{P}(\zeta_{s+1}=k \mid \zeta_{s}=i, \pi_{ij}^*, u^d=j)
\end{equation}
and let $p_{\pi_d}$ be the associated distribution over trajectories induced by some discrete state feedback policy, along with the discrete state action cost $c_d(u^d=j, \eta=i) = J_{ij}^*$ we may write the high level problem:
\begin{align}
    \pi_d^* &= \min_{\pi_d} J_d(\pi, \eta_0) \\
    &= \mathbb{E}p_{\pi_d}[\sum_{s=0}^{S} c_d(\eta_s, u^d_s)]
\end{align}
Our approximate control law is then given by $ \pi_{ij}^* \circ \pi_d^* \circ id(x) $

\subsection{Offline Low Level Problems: Linear Quadratic Regulator (LQR)}
\label{LQR}
Rather than solve the first-exit problem directly, we formulate an approximate problem by finding trajectories that end at specific `control priors' (see \ref{control-priors}).
Recall the low level problem given by:
\begin{align}
    \pi_{ij}(x_0) &= \arg \min_{\pi, S} J_{ij}(\pi, x_0, S) \\
    J_{ij}(\pi, x_0, S)&= \mathbb{E}_{\pi, x_0}[\sum_{t=0}^{S}c(x_t, u_t)]\\
    & \text{s.t. } (x_t, u_t) \in H_i \\
    & \text{s.t. } c(x, u) = 0 \text{ when } (x,u) \in \partial H_{ij}
\end{align}
In order to approximate this problem with one solvable by a finite horizon LQR controller, we adopt a fixed goal state, $x^* \in H_j$. Imposing costs $c_t(x_t, u_t) = u_t^T R u_t$ and $c_S(x_S, u_S) = (x - x^*) Q_f (x - x^*)$. Formally we solve,
\begin{align}
    \pi_{ij}(x_0) &= \arg \min_{\pi, S} J_{ij}(\pi, x_0, S) \\
    J_{ij}(\pi, x_0, S)&= \mathbb{E}_{\pi, x_0}[(x_S - x^*)^T Q_f (x_S - x^*) + \sum_{t=0}^{S-1} u_t^T R u_t]\\
\end{align}
by integrating the discrete Ricatti equation backwards. Numerically, we found optimising over different time horizons made little difference to the solution, so we opted to instead specify a fixed horizon (hyperparameter). These solutions are recomputed offline every time the linear system matrices change.

\textbf{Designing the cost matrices}
Instead of imposing the state constraints explicitly, we record a high cost which informs the discrete controller to avoid them. In order to approximate the constrained input we choose a suitably large control cost $R=rI$. We adopted this approach for the sake of simplicity, potentially accepting a good deal of sub-optimality. However, we believe more involved methods for solving input constrained LQR could be used in future, e.g. \cite{Bemporad2000}, especially because we compute these solutions offline.

\subsection{Online high level problem}
\label{online-problem}
The high level problem is a discrete MDP with a `known' model, so the usual RL techniques (approximate dynamic programming, policy iteration) apply. Here, however we choose to use a model-based algorithm with a receding horizon inspired by Active Inference, allowing us to easily incorporate exploration bonuses.

Let the Bayesian MDP be given by $\mathcal{M}_B = (S, A, P_a, R, P_\theta)$ be the MDP, where $p_a(s_{t+1}\mid s_t, a_t, \theta)\sim Cat(\theta_{as})$ and $p(\theta_{as}) \sim Dir(\alpha)$ 
We estimate the open loop reward plus optimistic information theoretic exploration bonuses 

\textbf{Active Inference conversion}
We adopt the Active Inference framework for dealing with exploration. Accordingly we adopt the notation $\ln \tilde{p}(s_{t}, a_t) = R(s_t, a_t)$ and refer to this 
`distribution' as the goal prior \cite{millidge2020relationship}, and optimise over open loop policies $\pi = (a_0, ..., a_T)$.

\begin{equation}
    J(a_{1:T}, s_0) = \mathbb{E}[\sum_{t=0}^{T} R(s_t, a_t) + IG_{p} + IG_{s}  \mid s_0, {a_{1:T}}]
\end{equation}
where parameter information-gain is given by $IG_{p} = D_{KL}[p_{t+1}(\theta) \mid \mid p_t(\theta)]$, with $p_t(\theta) = p(\theta \mid s_{0:t})$. In other words, we add a bonus when we expect the posterior to diverge from the prior, which is exactly the transitions we have observed least \cite{heins2022pymdp}.

We also have a state information-gain term, $IG_s = D_{KL}[p_{t+1}(s_{t+1}) \mid \mid p_t(s_{t+1})]$. In this case (fully observed), $p_{t+1}(s_{t+1}) = \delta_s$ is a one-hot vector. Leaving the term $\mathbb{E}_{t}[-\ln p_t(s_{t+1})]$ leading to a maximum entropy term \cite{heins2022pymdp}.

We calculate the above with Monte Carlo sampling which is possible due to the relatively small number of modes. Local approximations such as Monte Carlo Tree Search could easily be integrated in order to scale up to more realistic problems. Alternatively, for relatively stationary environments we could instead adopt approximate dynamic programming methods for more habitual actions. 

\subsection{Extracting the adjacency matrix from rSLDS}
\label{extract-adj}
In order to generate the possible transitions from the rSLDS, we calculate the set of active constraints for each region from the softmax representation, $p(z\mid x) = \sigma(Wx +  b)$. Specifically to check region $i$ is adjacent to region $j$ we verify the solution linear program:
\begin{align}
    - b_j  = &\min (W_i - W_j ) x \\
    & \text{s.t. } (W_i - W_k)x \leq (b_i - b_k) \text{ }\forall k \in [K] \\
    & \text{s.t. } x \in (x_{lb}, x_{ub})
\end{align}
Where $(x_{lb}, x_{ub})$ are bounds chosen to reflect realistic values for the problem. This ensures we only lift transitions to the discrete model, if they are possible. Again, these can be calculated offline. 

We initialise the entries of the transition model in the discrete MDP for possible transitions to 0.9 facilitating guided-exploration via information-gain through a count-based updates to the transition priors.

\subsection{Generating continuous control priors}
\label{control-priors}
In order to generate control priors for the LQR controller which correspond to each of the discrete states we must find a continuous state $x_i$ which maximises the probability of being in a desired $z$:

\begin{align}
x_i = \underset{x}{\arg\max} \, P(z=i|x, u)
\end{align}

For this we perform a numerical optimisation in order to maximise this probability. Consider that this probability distribution $P(z = i |x)$  is a softmax function for the i-th class is defined as:

\begin{align}
\sigma(v_i) = \frac{\exp (v_i)}{\sum_j \exp (v_j)}, v_i = w_i \cdot x + r_i
\end{align}

where $w_i$ is the i-th row of the weight matrix, $x$ is the input and $r_i$ is the i-th bias term. The update function used in the gradient descent optimisation can be described as follows:

\begin{align}
x \leftarrow x + \eta \nabla_x \sigma(v_i)
\end{align}

where $\eta$ is the learning rate and the gradient of the softmax function with respect to the input vector $x$ is given by:

\begin{align}
\nabla_x \sigma(v_i) = \frac{\partial \sigma(v_i)}{\partial v} \cdot \frac{\partial v}{\partial x} = \sigma(v_i)(\mathbf e_i - \sigma(v))\cdot W
\end{align}

in which $\sigma(v)$ is the vector of softmax probabilities, and $\mathbf e_i$ is the standard basis vector with 1 in the i-th position and 0 elsewhere. The gradient descent process continues until the probability $P(z=i | x)$ exceeds a specified threshold $\theta$ which we set to be 0.7. This threshold enforces a stopping criterion which is required for the cases in which the region $z$ is unbounded.
\clearpage

\subsection{Model-free RL baselines}
\subsubsection{Soft-Actor Critic with 2 Q-functions }

\begin{table}[h]
\caption{Summary of the Soft Actor-Critic algorithm with multiple Q-functions.}
\label{SAC-table}
\vskip 0.15in
\begin{center}
\begin{small}
\begin{sc}
\begin{tabular}{lcccr}
\toprule
Component & Input \\
\midrule
Q-network    & 3×256×256×256×2\\
Policy network & 2×256×256×256×2\\
Entropy regularization coeff    & 0.2\\
Learning rates (Qnet + Polnet)    & 3e-4\\
Batchsize     & 60\\
\bottomrule
\end{tabular}
\end{sc}
\end{small}
\end{center}
\vskip -0.1in
\end{table}
\subsubsection{Actor-Critic}

\begin{table}[h]
\caption{Summary of the Actor-Critic algorithm}
\label{AC-table}
\vskip 0.15in
\begin{center}
\begin{small}
\begin{sc}
\begin{tabular}{lcccr}
\toprule
Component & Input \\
\midrule
Feature Processing    & StandardScaler, RBF Kernels (4 $\times$ 100) \\
Value-network    & 4001 parameters (1 dense layer)\\
Policy network & 802 parameters (2 dense layers) \\
Gamma & 0.95\\
Lambda    & 1e-5\\
Learning rates (Policy + Value)    & 0.01\\
\bottomrule
\end{tabular}
\end{sc}
\end{small}
\end{center}
\vskip -0.1in
\end{table}

\subsection{Model-based RL baseline}
\label{DQNMBE}
\subsubsection{a Deep Q-Network with Model-based Exploration (DQN-MBE)}

\begin{figure}[htbp]
    \centering
    \subfigure[HHA (our model) on Continuous Mountain Car]{
        \includegraphics[width=0.4\linewidth]{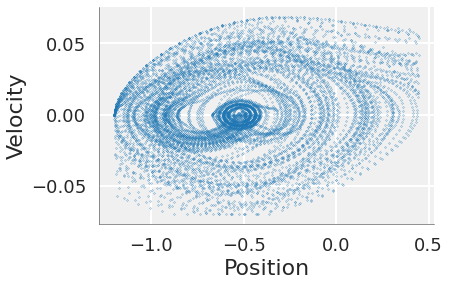}
        \label{fig:HHA_IG}
    }
    \hfill
    \subfigure[DQN-MBE on Discrete Mountain Car]{
        \includegraphics[width=0.4\linewidth]{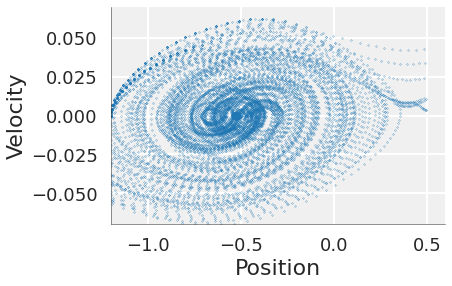}
        \label{fig:DQN-MBE}
    }
    \caption{\textbf{On Continuous Mountain Car, our model (HHA) competes with the state-space coverage achieved by model-based baselines on Discrete Mountain Car (an easier problem)} State-space coverage after 10,000 timesteps on (a) Continuous Mountain Car task using our model (HHA) and (b) Discrete Mountain Car task using a Deep Q-Network with Model-Based Exploration (DQN-MBE) \cite{Gou2019}. Exact parameters in Table~\ref{dqn-table}.}
    \label{fig:combined}
\end{figure}

\begin{table}[H]
\caption{Summary of DQN-MBE algorithm \cite{Gou2019}}
\label{dqn-table}
\vskip 0.15in
\begin{center}
\begin{small}
\begin{sc}
\begin{tabular}{lcccr}
\toprule
Component & Input \\
\midrule
Q-network    & 1 hidden-layer, 48 units, ReLU \\
Dynamics Predictor Network (Fully Connected) & 2 hidden-layers (each 24 Units), ReLU \\
$\epsilon$ minimum    & 0.01 \\
$\epsilon$ decay    & 0.9995\\
Reward discount & 0.99 \\
Learning rates (Qnet / Dynamics-net)    & 0.05 / 0.02\\
Target Q-network update interval & 8\\
Initial exploration only steps     & 10000\\
Minibatch size (Q-network)   & 16\\
Minibatch size (dynamics predictor network)     & 64\\
Number of recent states to fit probability model     & 50\\
\bottomrule
\end{tabular}
\end{sc}
\end{small}
\end{center}
\vskip -0.1in
\end{table}

\end{document}